\documentclass[11pt,twocolumn]{scrartcl}

\usepackage{casa_conf}	
\usepackage{graphicx}	
\usepackage{adjustbox}
\usepackage{amsmath}
\usepackage{pifont}
\usepackage{amssymb}
\usepackage{subcaption}
\usepackage{svg}
\usepackage{hyperref}
\usepackage{booktabs}
\usepackage{orcidlink}

\title{XmoPipe: A Pipeline for Large-Scale In-the-Wild Human Motion Dataset Construction}

\author{
    Nathan Salazar$^{1}$\,\orcidlink{0000-0001-8581-0536} \and
    Emmanuel Dellandréa$^{1}$ \orcidlink{0000-0001-7346-228X} \and
    Mathieu Lefort$^{2}$ \orcidlink{0000-0001-8581-0536} \and
    Alexandre Meyer$^{3}$ \orcidlink{0000-0002-0249-1048}
}

\date{
\small
$^{1}$ Ecole Centrale de Lyon, CNRS, INSA Lyon, Universite Claude Bernard Lyon 1, LIRIS, UMR5205, 69130 Ecully, France\\
\texttt{\{nathan.salazar, emmanuel.dellandrea\}@ec-lyon.fr}\\
$^{2}$ Univ Rennes, Inria, CNRS, IRISA - UMR 6074; F-35000 Rennes, France\\
\texttt{mathieu.lefort@univ-rennes.fr}\\
$^{3}$ Universite Claude Bernard Lyon 1, CNRS, INSA Lyon, LIRIS, UMR5205, 69622 Villeurbanne, France\\
\texttt{alexandre.meyer@univ-lyon1.fr}
}

\begin{document}

\maketitle

\begin{abstract}
Large-scale human motion datasets are essential for training robust motion models for analysis, synthesis, and understanding. While marker-based motion capture provides precise data, it is costly and limited in scale and diversity. Recent advances in monocular motion capture and video-language understanding open the way to extract plausible motion from unconstrained online videos.
We present a scalable pipeline for constructing \emph{in-the-wild} human motion datasets. From a few keywords, the system retrieves videos, extracts 3D body and facial motion, and generates high-level textual descriptions. The pipeline is flexible, enabling targeted collection of various motions, multi-person interactions, or expressive behaviors. We demonstrate its quality by training motion reconstruction and motion generation models, showing performance comparable to models trained on traditional motion capture datasets and strong cross-dataset generalization.\\ Code and Motion Data are available at \href{https://github.com/NatSalaz/xmopipe}{https://github.com/NatSalaz/xmopipe}.
\end{abstract}
\linebreak
\linebreak

\section{Introduction}
\label{sec:intro}
Human motion is a central topic in computer vision, graphics, and robotics.
Accurately modeling how humans move is essential for tasks such as data-driven animation systems, human–computer interaction, virtual avatar control, and activity recognition. However, human motion is inherently complex and highly diverse. It spans a wide range of activities, interaction patterns, body shapes, expressions, and styles, and is strongly influenced by context. Capturing this diversity in a single dataset remains a major challenge, but it is a prerequisite for learning robust and generalizable motion models.

Early human motion datasets were primarily built using marker-based, multi-camera motion capture systems. These setups provide highly accurate 3D joint positions and have played a crucial role in advancing motion modeling and synthesis. Datasets obtained through such systems remain a gold standard in terms of geometric precision and temporal consistency. Nevertheless, constructing large motion-capture-based datasets is expensive, time-consuming, and logistically complex. As a result, most existing datasets are limited in scale and variety, and often rely on a small number of actors performing scripted actions in controlled environments. While these datasets are invaluable, they do not fully reflect the richness and variability of human motion as it occurs in everyday life.

Ideally, human motion datasets should be collected in the wild, capturing natural behaviors, spontaneous interactions, and a broad spectrum of activities across diverse environments and persons. 
Achieving this level of realism has long been challenging due to the lack of reliable motion capture methods outside laboratory settings.

Recent advances in monocular motion capture and video analysis have significantly changed this, with modern methods such as GVHMR\cite{shen2024gvhmr}, TRAM\cite{wang2024tramglobaltrajectorymotion} or MultiHMR\cite{baradel2024multihmrmultipersonwholebodyhuman} now capable of recovering full-body motion, some including facial dynamics, directly from monocular videos with increasing accuracy. These developments make it possible to envision leveraging large-scale video repositories, such as YouTube, as a source of diverse and realistic human motion data. In parallel, recent progress in video-based large language models such as ChatGPT-4o\cite{openai2024gpt4ocard}, Video-LLama\cite{zhang2023videollamainstructiontunedaudiovisuallanguage}, Qwen-VL\cite{Qwen-VL}  has enabled automatic semantic understanding of video content. These models can generate textual descriptions of scenes, actions, and interactions, providing rich high-level annotations that were previously costly to obtain at scale.

In this work, we build upon these advances to propose XmoPipe, a flexible and scalable pipeline for automatically constructing large-scale, in-the-wild human motion datasets from online videos. Starting from a small set of user-defined keywords, XmoPipe automatically augments video queries, retrieves relevant content, and extracts human motion—covering both body and facial dynamics—using monocular motion capture techniques. Video-language models are then employed to produce textual descriptions of the observed actions and behaviors.
Thanks to its fully customizable design, XmoPipe can be easily adapted to different dataset objectives or motion domains—such as sports activities, multi-person interactions, or specific movement categories—simply by modifying the initial keyword prompts. 

In addition to releasing the pipeline as a contribution, we provide a substantial dataset that is large-scale and highly diverse, capturing multi-person motions across a wide range of actions in unconstrained, real-world environments. We first validate the dataset via a motion reconstruction task: combining our dataset with traditional MoCap data improves reconstruction accuracy and demonstrates the benefits of cross-dataset augmentation for generalization. We also train a state-of-the-art Motion Diffusion Model (MDM) for text-to-motion generation. The resulting motions maintain quality comparable to models trained on MoCap-only datasets, while offering richer and more diverse textual annotations that support a wider range of actions and more flexible prompts.



This paper is organized as follows. The \autoref{sec:relworks} reviews related work on motion capture, in-the-wild datasets, and video-language models. \autoref{sec:xmo} presents the XmoPipe pipeline, detailing video collection, motion extraction, facial integration, and semantic enrichment. \autoref{sec:xmodata} describes the resulting dataset and its statistical properties. \autoref{sec:val} evaluates the dataset on reconstruction and generation tasks. \autoref{sec:conclusion} concludes and outlines future directions.

\section{Related Work}
\label{sec:relworks}
\textbf{Human Motion Datasets.} Large-scale human motion datasets are typically captured in controlled environments using marker-based motion capture (MoCap) systems, ensuring highly accurate 3D poses. Human3.6M \cite{Human36M} and AMASS \cite{AMASS:ICCV:2019} are representative single-person MoCap datasets: Human3.6M was primarily designed for 3D pose estimation and action recognition, providing multi-view recordings within reconstructed indoor environments, while AMASS aggregates multiple MoCap datasets into a unified representation, making it widely used for motion modeling and generation.

More recent datasets extend this paradigm to multimodal learning. HumanML3D \cite{guo2022humanml3d} augments single-person motions with natural language descriptions for text-to-motion generation, whereas InterHuman/InterGen \cite{liang2024intergen} focuses on two-person interactive motions with textual annotations, enabling conditional generation of coordinated human interactions. Despite their high fidelity, scaling such datasets to broader diversity and interaction complexity remains challenging.

\begin{figure*}[!t]
    \centering
    \includegraphics[width=1.0\textwidth]{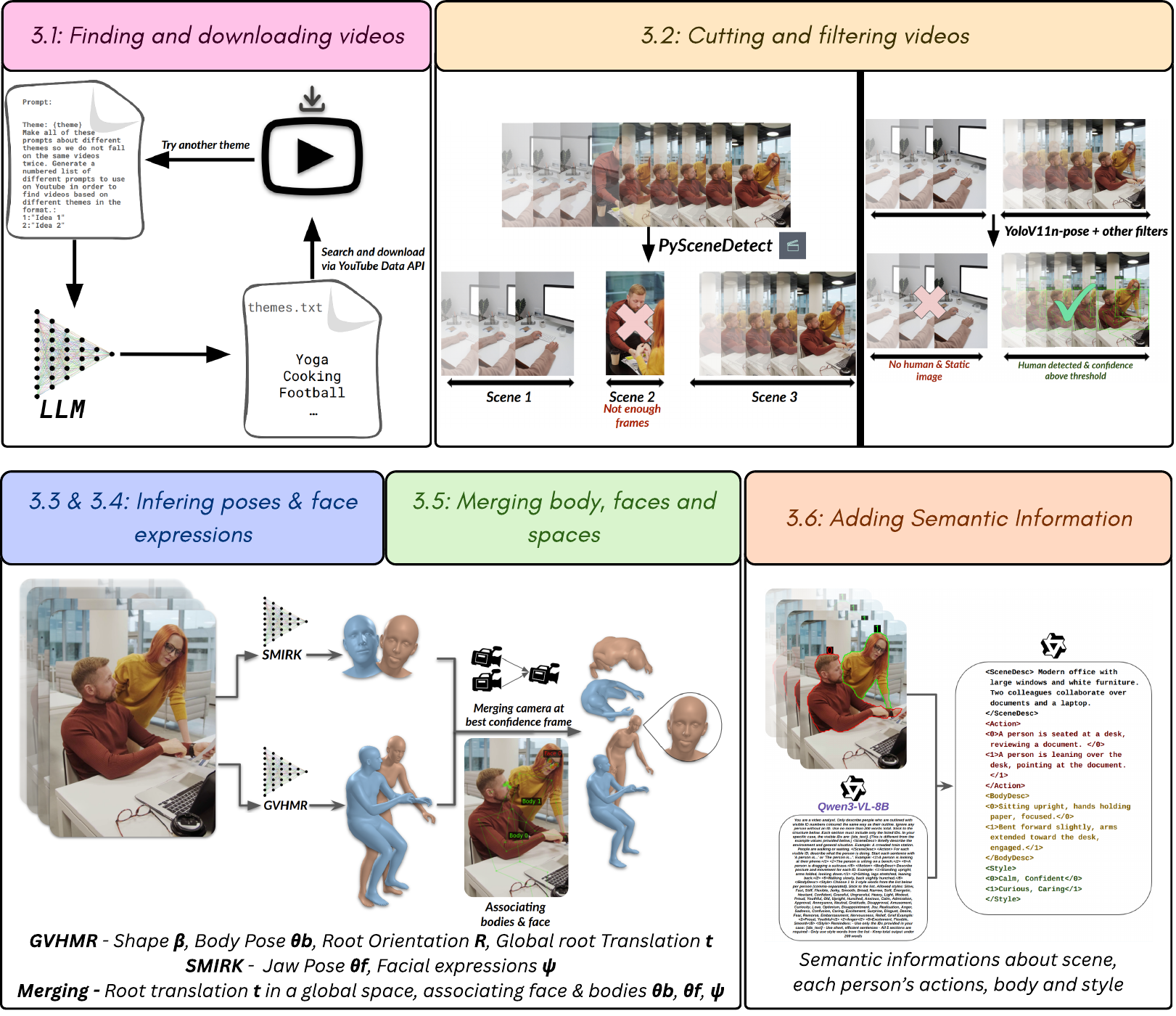}
    \caption{XmoPipe: from a few augmented initial keywords (\ref{subsec:videos}), raw videos are filtered and segmented(\ref{subsec:segment}), once relevant, they are transformed into multi-person SMPL-X motion data (\ref{subsec:inferposes},\ref{subsec:inferfaces},\ref{subsec:merge}) with aligned textual scene descriptions (\ref{subsec:semantic}).}
    \label{automaton}
\end{figure*}

Recently, large-scale motion datasets constructed from in-the-wild online videos have emerged. In this paper, we use the term “pseudo-ground-truth” (pseudo-GT) to refer to 3D human motion obtained via reconstruction from monocular videos, in contrast to traditional motion capture (MoCap) data, which is recorded using marker-based systems. For example, Motion-X \cite{lin2023motionx} combines internet-sourced videos with curated motion capture data to increase diversity, while Go-to-Zero \cite{fan2025zerozeroshotmotiongeneration} leverages large quantities of video-derived motion to improve zero-shot generalization. These works primarily emphasize dataset scale and performance gains.

In contrast, our contribution spans both the release of a large-scale motion dataset and the design of an open, flexible construction pipeline. Our pipeline is modular and fully open-source, enabling reproducible data collection and motion extraction from raw web videos. Unlike prior approaches, where processing pipelines are often only partially released or lack certain components, our framework provides a complete and reproducible implementation. Our codebase allows researchers to customize video retrieval by modifying keyword selection, as well as to adapt the motion reconstruction and annotation components with limited additional effort. This design makes it straightforward to expand existing datasets or to build new, domain-specific motion corpora—such as sports- or art-focused datasets—without requiring motion capture hardware.

\textbf{Monocular Motion Capture from Video.} Monocular motion capture methods now enable 3D human pose and motion recovery directly from RGB videos, often using parametric body models such as SMPL-X \cite{SMPL-X:2019}. HMR\cite{kanazawa2018hmr} introduced single-image pose and shape estimation, VIBE\cite{kocabas2020vibe} improved temporal consistency, and WHAM\cite{shin2024wham} further improved robustness, long-term motion recovery, and expressive detail. In our work, we leverage GVHMR\cite{shen2024gvhmr} as it delivers highly realistic and physically plausible motions in a global coordinate space, enabling accurate large-scale \emph{in-the-wild} motion extraction without specialized capture setups. This step is detailed further in \autoref{subsec:inferposes}.

\textbf{Video Understanding and Video-Language Models.} Video-language models provide high-level semantic descriptions of video content. VideoBERT~\cite{sun2019videobert} pioneered joint video-text representation learning, Flamingo~\cite{alayrac2022flamingo} extended few-shot capabilities to video, and Video-LLaMA~\cite{zhang2023videollama} supports instruction-guided video understanding. Such models enable automatic and scalable labeling of large online video collections for action and behavior annotation. This is particularly valuable given that manual annotation requires human workers to watch videos and produce textual descriptions, which is time-consuming, costly, and difficult to scale for large datasets.

\textbf{Motion generation models.} Recent state-of-the-art approaches for 3D human motion generation rely on reconstruction-based latent models such as VAEs and VQ-VAEs to learn structured motion priors before generative modeling. ACTOR\cite{petrovich2021actionconditioned3dhumanmotion} introduced an action-conditioned Transformer-VAE framework to encode and reconstruct motion sequences in a continuous latent space. Later works such as T2M-GPT\cite{zhang2023t2mgptgeneratinghumanmotion}  adopted vector-quantized representations to discretize motion tokens and improve compositionality and stability. Building upon these learned priors, diffusion-based models such as MDM\cite{tevet2022humanmotiondiffusionmodel} generate motions through iterative denoising, while latent diffusion approaches like LDM\cite{chen2023executingcommandsmotiondiffusion} and MoMaDiff\cite{momadiff} perform diffusion in a compressed motion latent space, improving efficiency, realism, and semantic alignment compared to pure VAE-based generation.

\section{The XmoPipe pipeline }
\label{sec:xmo}
As described in \autoref{automaton}, XmoPipe is an end-to-end pipeline that starts with a few keywords chosen by the user to automatically collect raw videos used to extract multi-person motion capture data, enriched with structured textual descriptions of the observed scene and interactions. It first relies on a large language model to generate diverse search queries and retrieve Creative Commons videos from YouTube, as explained in \autoref{subsec:videos}. Then, as shown in \autoref{subsec:segment}, videos are segmented into continuous shots and filtered to retain only dynamic segments containing visible humans. For each valid sequence for each individual, body motions (\autoref{subsec:inferposes}) and facial meshes (\autoref{subsec:inferfaces}) are reconstructed and merged into a unified SMPL-X representation (\autoref{subsec:merge}); their relative positions are computed. Finally, semantic captions are generated to describe the scene as well as the motion of each individual, as detailed in \autoref{subsec:semantic}.

\subsection{Finding and Downloading Videos}
\label{subsec:videos}
The first step is to identify relevant videos containing rich and expressive human interactions. 
Starting from a small set of given initial keywords, we propose using a large language model, Qwen3-4B\cite{Qwen-VL} to automatically generate search queries with a large lexical scope. The use of an LLM enables the automatic generation of semantically diverse and context-aware queries, covering a wider range of human interactions than hand-crafted keywords, thus increasing the diversity and expressiveness of the collected dataset. These queries are used to collect videos from YouTube, while ensuring that only Creative Commons-licensed videos are retained, thanks to filtering performed with the YouTube Data API v3\cite{google_youtube_api_v3}.


\subsection{Cutting and Filtering Videos}
\label{subsec:segment}
The videos are then segmented in a first pass using SceneDetect\cite{PySceneDetect}, which identifies changes in shots to cut the recordings into visually coherent scenes. This step helps isolate the sequences of interest for the detection and reconstruction stages. 
To further filter and resegment relevant passages, a second pass is applied based on the real-time 2D human pose detection model YoloV11n-pose \cite{yolo11_ultralytics}, which provides a fast and lightweight estimate of human presence. This step is only intended to discard irrelevant or empty segments; accurate 3D motion capture is performed later(\autoref{subsec:inferposes}). Only segments containing at least one individual detected with an average confidence score over the segment above a predefined threshold are retained. If all the confidence scores of a frame are too low, we stop the segment.


A tracking module is then used to group together the appearances of the same individual throughout the sequence. In the event of discontinuity or significant changes in appearance between frames, a new sequence is created to avoid any ambiguity in subsequent reconstructions. 
This sequencing ensures that only scenes where individuals are continuously present and clearly visible are processed. To filter out static content, we also compute a normalized optical flow magnitude within the bounding boxes of detected individuals using the Farnebäck method\cite{opticalflow}. Segments with a normalized optical flow magnitude below an empirically determined threshold are considered static and discarded.

All the predefined values are available in the appendix.
\subsection{Inferring pose}
\label{subsec:inferposes}
The goal of this step is to recover accurate global body pose parameters for each individual in the video.
After evaluating several existing approaches, including WHAM \cite{shin2024wham} and MultiHMR \cite{baradel2024multihmrmultipersonwholebodyhuman}, we selected the GVHMR\cite{shen2024gvhmr} model for global body pose estimation. This choice is motivated by its strong performance both quantitatively and qualitatively in recovering global human movement from monocular videos, its explicit modeling of contact constraints to reduce foot sliding artifacts, and its robustness in dynamic scenes from the real world.

GVHMR outputs only body parameters in the SMPL \cite{SMPL:2015} format, a widely used parametric human body model. These parameters are embedded into the standard SMPL-X \cite{SMPL-X:2019} format, which also includes face and hand components. In our pipeline, body and face are reconstructed (\ref{subsec:inferposes},\ref{subsec:inferfaces}), while hand reconstruction is left for future work. Using the SMPL-X representation facilitates seamless integration with extended models while maintaining accurate body-centric pose estimation, despite GVHMR’s predictions being restricted to the SMPL body joint set.

As part of the GVHMR pipeline, person detection and 2D keypoints extraction are performed using the ViTPose-h model \cite{xu2022vitpose}. We save and reuse the 2D keypoints for later merging with faces \ref{subsec:merge}.
GVHMR is originally designed for single-person motion reconstruction. We extend its use to multi-person scenes by applying the model independently to each tracked individual. Only detected persons who meet our confidence and temporal consistency thresholds are processed, ensuring robust reconstruction without introducing inter-person coupling at this stage.
In addition to 2D keypoint locations, ViTPose-h provides per-frame detection confidence scores for each individual. We exploit these scores as a lightweight indicator of pose reliability. Frames whose confidence falls below a predefined threshold (reported in Appendix \ref{sec:annex}) are flagged as unreliable. Such frames typically correspond to occlusions, motion blur, extreme poses, or partial detections, and are more likely to yield degraded 3D pose estimates. These confidence flags are preserved for downstream processing and are later used to define robust temporal boundaries and to exclude unstable frames from smoothing and normalization steps.

\subsection{Inferring face expressions}
\label{subsec:inferfaces}
In addition to body reconstruction, our pipeline includes a module dedicated to facial reconstruction, with two main objectives: estimating facial expression and jaw-related SMPL-X parameters for each detected individual, and ensuring their consistent alignment and integration with the reconstructed body motion.
Face detection is performed using YOLOv8l \cite{7780460}, applied independently to each video frame. For each detected face, 2D facial landmarks are extracted using MediaPipe \cite{lugaresi2019mediapipeframeworkbuildingperception}, which provides accurate localization of key facial landmarks. These two models are simple and computationally lightweight for large-scale processing, they also empirically provide good performances.
The extracted 2D facial landmarks are used as input to the SMIRK model\cite{retsinas20243dfacialexpressionsanalysisbyneuralsynthesis}, which predicts facial expression and jaw motion parameters from monocular video in a representation compatible with SMPL-X.
This model was chosen for its superior performance compared to alternative models such as EMOCA\cite{EMOCA:CVPR:2021}, particularly in terms of expression fidelity in videos under natural conditions.

\subsection{Data Fusion and Smoothing}
\label{subsec:merge}
To associate faces with reconstructed bodies, we directly match their 2D keypoints in image space.
For each frame, we compare the pixel coordinates of the nose and eye landmarks detected by MediaPipe(\ref{subsec:inferfaces}) with the corresponding nose and eye keypoints predicted by ViTPose-h(\ref{subsec:inferposes}).
The Euclidean distance between these landmarks is used as a matching cost. We then solve the resulting assignment problem using the Hungarian algorithm, ensuring a one-to-one correspondence between faces and body instances.

The SMPL-X body parameters predicted by GVHMR are augmented with the facial expression and jaw parameters estimated by SMIRK, resulting in a unified SMPL-X representation per individual. Temporal smoothing is applied to the body relative joints positions and facial expression parameters using a one-dimensional Gaussian filter. This temporal filtering reduces high-frequency jitter and visual tremors caused by minor detection inconsistencies, while preserving short-term expression dynamics.

Isolated facial detections are removed based on an empirical temporal criterion. Short facial tracks lasting fewer than a prefixed number of frames are discarded, as they typically correspond to unstable detections.
All processing is performed in an offline setting, as the employed models are inherently designed for offline inference and do not impose real-time constraints
.
To further improve motion quality, frames flagged with low confidence scores at the beginning and end of each character sequence are removed, as such artifacts often result from imperfect temporal segmentation of the original videos. 
GVHMR provides both camera-relative and global positions. We align and unify the global coordinate spaces by estimating translation offsets in camera space, using the most reliable frames where individuals are clearly visible and ViTPose detections have the highest confidence, ensuring stable alignment across sequences.
Finally, the global coordinate system is normalized by centering the scene at the origin, with the first individual’s root in the initial frame positioned at origin.

Once again all the hyperparameter values are available in the appendix and can be easily modified in our code.
\subsection{Adding Semantic Information}
\label{subsec:semantic}
Semantic information provides additional context that helps understanding actions, interactions, and identities, which is important for many applications and whose benefits we illustrate in the Results section. To this end, we integrate semantic descriptions automatically extracted from videos using video captioning models. 
We chose Qwen3-VL-8B \cite{Qwen3VL} to generate global visual descriptions, as it is a state-of-the-art, well-established model that produces high-quality captions.
Inspired by the method proposed in \cite{zhang2024visualpromptingllmsenhancing}, we also explore the use of modified videos to guide the model towards relevant individuals. 
More specifically, we use the bounding boxes around target individuals computed for GVHMR inference to crop the corresponding image regions, which are then outlined using FastSAM \cite{zhao2023fast}, thereby facilitating caption generation with Qwen-VL.
These enriched annotations allow us to capture important contextual elements, which are particularly useful for disentangling multi-person sequences.
The semantic data is organized into four sub-parts: 
(i) a global scene description (one per video),
(ii) per-individual appearance descriptions,
(iii) action descriptions, and
(iv) style descriptions.
 providing text–motion alignment
For the style description, we injected a controlled list of affective and movement-related descriptors into the prompt to guide the Vision-Language Model toward captions aligned with a predefined expressive space, inspired by GoEmotions \cite{GoEmotions} and Laban Movement Analysis \cite{laban1950mastery}.

\section{XmoPipe Dataset}
\label{sec:xmodata}
In order to validate our pipeline, we automatically collect and process hundreds of hours of Creative Commons–licensed videos using our pipeline, resulting in 197 hours of filtered footage. This generic dataset will allow us to validate the pipeline use on diverse types of motions from everyday life and sports. The videos are then segmented, yielding 172,182 individual sequences with a total duration of 312 hours, the sequences also are 92,182 multi/mono-person sequences totaling 180 hours.

\begin{figure}[ht]
    \centering
    \begin{subfigure}{\linewidth}
        \centering
        \includegraphics[width=0.48\columnwidth]{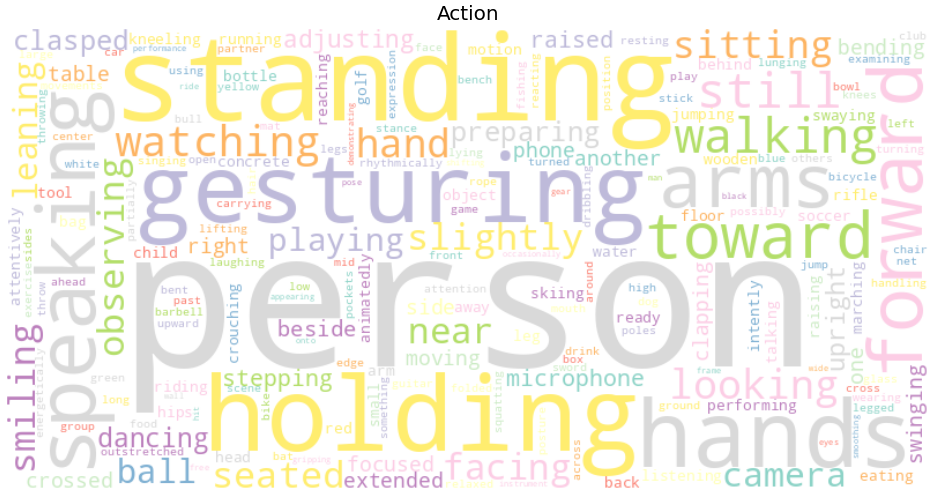}
        \includegraphics[width=0.48\columnwidth]{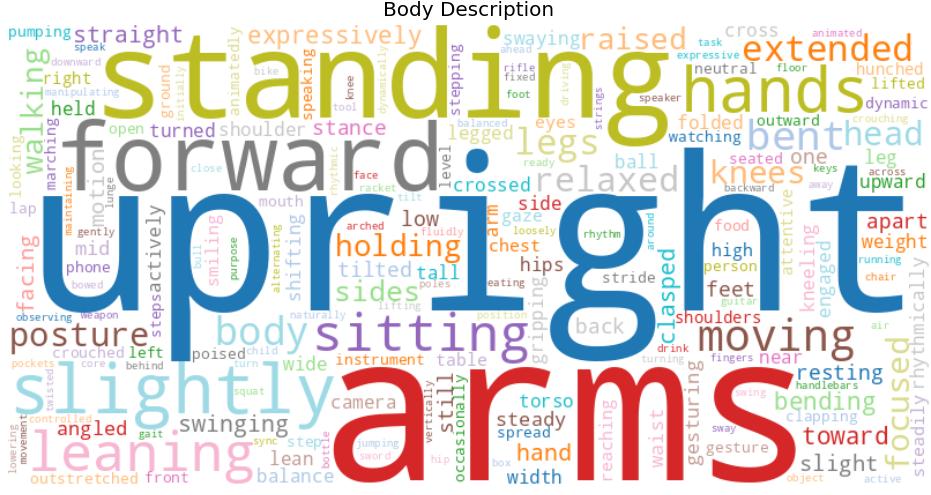}\\[2mm]
        \includegraphics[width=0.48\columnwidth]{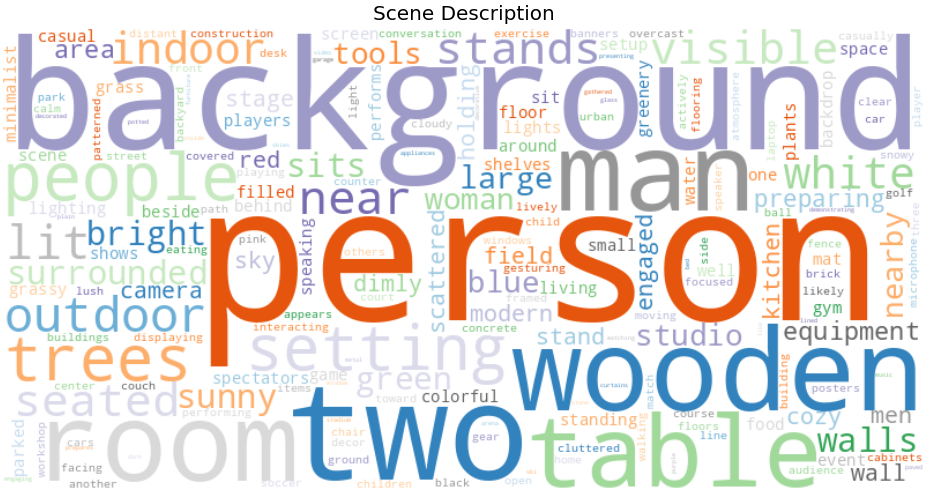}
        \includegraphics[width=0.48\columnwidth]{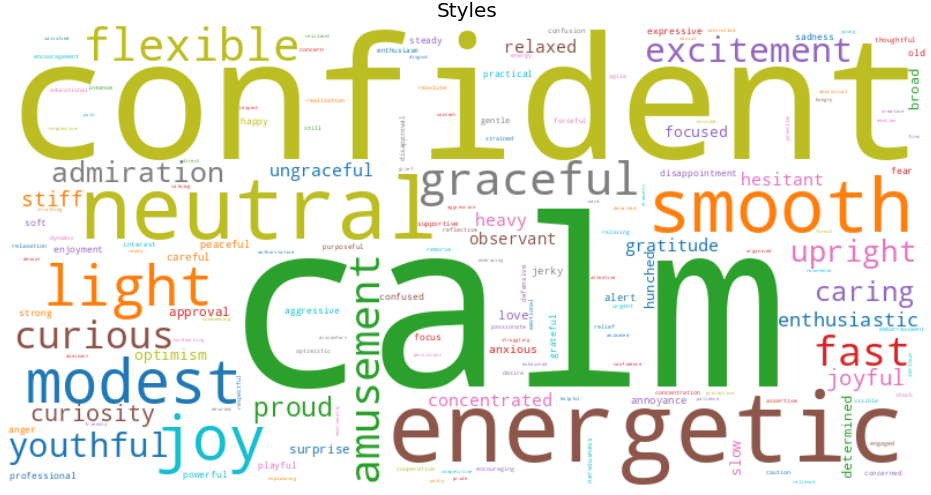}
        \caption{\footnotesize Wordclouds computed from word occurrences in textual descriptions. Respectively Action, Body description, Scene Description and Style Description sections.}
        \label{fig:wordcloud}
    \end{subfigure}
    \\[4mm]
    \begin{subfigure}{\linewidth}
        \centering
        \includegraphics[width=0.48\textwidth]{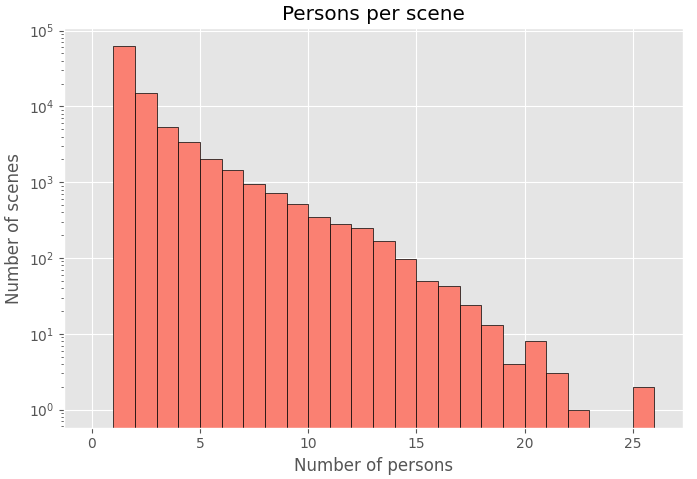}
        \caption{\footnotesize Number of persons per scene.}
        \label{fig:personperscene}
    \end{subfigure}
    \caption{Dataset statistics.}
    \label{fig:stats}
\end{figure}

A word cloud visualizations in \autoref{fig:wordcloud} provide an overview of the most frequent terms in the textual annotations, showing the distribution of the dataset vocabulary for each textual subdescription.
Together, these statistics illustrate the semantic, structural, and contextual diversity of XmoPipe, which underpins its potential for both reconstruction and generative motion modeling.
\autoref{fig:personperscene} show the distribution of individual per motion scene.


\section{Validation and Comparison} 
\label{sec:val}

\subsection{Existing Dataset}
\begin{table}[ht]
  \centering
  \setlength{\tabcolsep}{3pt}
  \renewcommand{\arraystretch}{0.95}
  \begin{adjustbox}{max width=0.47\textwidth}
    \begin{tabular}{@{}l c c c c c 
    c@{}}
      \toprule
      Dataset & \#Seq & \#Txt & Dur. & Parts & S/M 
      & Qual. \\
      \midrule
      KIT \cite{Plappert2016} 
        & 5.7k & 5.7k & 11.2h & B 
        & S 
        & MoCap \\

      HumanML3D \cite{guo2022humanml3d} 
        & 14.6k & 44.9k & 28.6h & B 
        & S 
        & MoCap \\      

      BEAT \cite{liu2022beatlargescalesemanticemotional}  
        & 2.5k & 2.5k & 76.0h & B,F,H 
        & S 
        & MoCap \\

      MotionX \cite{lin2023motionx}    
        & 81.1k & 142k & 144h & B,F,H 
        & S 
        & Both \\

      MotionLib \cite{wang2025scalinglargemotionmodels}                    
        & 1.21M & 2.48M & 1456h & B  
        & S \& M 
        & Pseudo-GT \\

      MotionMillion \cite{fan2025zerozeroshotmotiongeneration}                    
        & 2M & $>$40M & 2000h & B    
        & S \& M 
        & Pseudo-GT \\

      \textbf{Ours} 
        & 172k & 172k & 312h/180h & B,F  
        & S \& M (rel.) 
        & Pseudo-GT \\
      \bottomrule
    \end{tabular}
  \end{adjustbox}
  \caption{Comparison with SOTA motion-text datasets. B: Body, F: Face, H: Hands., S: Single, M:Multi-person,rel.:relative.
  Pseudo-Ground-Truth (pseudo-GT) refers to 3D human motion obtained via reconstruction from monocular videos. MoCap as traditional motion capture.
  }
  \label{comparison}
\end{table}


As shown in \autoref{comparison}, larger datasets such as MotionLib and MotionMillion contain more data but achieve this through aggressive filtering, retaining only very well-defined single-person actions. In contrast, our dataset preserves multi-person sequences with approximate relative positioning, which introduces additional challenges for motion extraction and annotation, while offering richer context and interactions. Moreover, it provides facial motion that is fully integrated with the body, along with detailed textual descriptions that go beyond actions to capture posture, motion style, and scene context.

We adopt a reconstruction-based evaluation as a controlled diagnostic task to assess the quality and consistency of the motion data produced by our pipeline. Reconstruction is not the final objective of our approach; instead, it provides a simple and well-understood setting that allows us to isolate the impact of dataset quality on learning.
By training the same reconstruction model on different datasets and evaluating both in-domain and cross-dataset performance, we can measure how well the learned motion representations generalize beyond their training distribution. This protocol is particularly appropriate for automatically collected, in-the-wild data, where some level of noise is inevitable but increased diversity is expected to improve robustness and generalization.

For the reconstruction evaluation, we rely on two complementary datasets: HumanML3D \cite{guo2022humanml3d} and IDEA400, a subset from MotionX \cite{lin2023motionx}. HumanML3D is used for motion-based evaluations, as it constitutes a widely adopted benchmark for assessing motion quality and diversity in human motion generation and understanding tasks. While HumanML3D and IDEA400 focus on motion quality under controlled or semi-controlled settings, our approach enables large-scale automatic collection from real-world videos.
HumanML3D is constituted of motion sequences from AMASS \cite{AMASS:ICCV:2019} and HumanAct12 \cite{humanact12}, comprising 14,616 motion sequences paired with 44,970 textual descriptions, with multiple annotations per motion.
To evaluate reconstruction quality in settings closer to our data collection pipeline, we use IDEA400 is a subset of MotionX \cite{lin2023motionx} and the largest component of MotionX outside of AMASS. IDEA400 contains 12,042 motion sequences derived from monocular human motion estimation applied to front-facing video recordings of actors performing everyday actions, with each sequence associated with a single textual description. Although it does not provide motion capture ground truth, IDEA400 offers structured and relatively clean pseudo-GT reconstructions obtained under controlled recording conditions.
Together, these two datasets enable evaluation across diverse motion distributions. As shown in \autoref{comparison}, XmoPipe complements these benchmarks by providing multi-person scenes, facial motion reconstruction, and relative spatial information, which are not available in HumanML3D and IDEA400.

\subsection{Quality Validation}
For the following sections, we convert all motions to the 263-dimensional HumanML3D representation \cite{guo2022humanml3d}. We adopt this representation even if mono-person centered as it is widely used in the community, relatively simple, and allows to conveniently evaluate and compare multiple text-to-motion models.
This representation encodes root motion, local joint positions, velocities, rotations, and foot-contact features.
In a coordinate system with $Y$ up vector, each pose $\mathbf{p}$ is defined as
\[
\mathbf{p} = (\dot{r}_a,\ \dot{r}_x,\ \dot{r}_z,\ r_y,\ \mathbf{j}_p,\ \mathbf{j}_v,\ \mathbf{j}_r,\ \mathbf{c}_f),
\]
where 
$\dot{r}_a \in \mathbb{R}$ is the root angular velocity around $Y$ axis;
$(\dot{r}_x,\dot{r}_z) \in \mathbb{R}^2$ are the root linear velocities in the plane (XZ);
$r_y \in \mathbb{R}$ is the root height;
$\mathbf{j}_p \in \mathbb{R}^{3j}$,  $\mathbf{j}_v \in \mathbb{R}^{3j}$, $\mathbf{j}_r \in \mathbb{R}^{6j}$ are the local joint position, velocity, and 6D rotation for each of the 21 local joints, excluding the root, which encodes only the velocity, following the HumanML3D convention;
and $\mathbf{c}_f \in \mathbb{R}^4$ denotes the foot–contact features.


\subsubsection{Reconstruction task}
To evaluate out-of-distribution motion reconstruction and study the effect of dataset diversity on generalization, we conduct two experiments. First, we adopt the motion reconstruction model of MoMaDiff \cite{momadiff}, a motion VAE architecture, to serve as a standard baseline for evaluating reconstruction performance. Second, we train a Motion Diffusion Model (MDM) \cite{tevet2022humanmotiondiffusionmodel} with text–motion alignment. 
The reconstruction task serves as a controlled baseline to confirm that our dataset can achieve performance on par with HumanML3D under a simple autoencoding setup. While no substantial improvements are expected here, the benefits of our large-scale, diverse data become apparent when training higher-capacity generative models like MDM for text-to-motion generation.

For the reconstruction task based on MoMaDiff's work, we kept only their autoencoding parts and ignored the original end-to-end objectives. MoMaDiff uses a sequential variational autoencoder with a continuous latent space, motion sequences are encoded into a temporally downsampled representation. Encoder and decoder are CNNs with ResNet blocks using a downsampling factor of 8 and a latent dimension of $8 \times 256$. We train this backbone on XmoPipe, HumanML3D, and Idea400, as well as on various combinations of these datasets, using a learning rate of $1 \times 10^{-5}$, a batch size of 256, and motion sequences of size $[64,263]$. To ensure robust evaluation, we perform cross-dataset training and testing, effectively implementing a leave-one-dataset-out scheme, and report reconstruction errors on each dataset in turn. This protocol allows us to study not only in-domain reconstruction but also cross-dataset generalization, allowing us to analyze the impact of dataset diversity and scale. Hyperparameters are given in appendix, we followed MoMaDiff parameters outside the latent dimension, we lowered it from 512 to 256 in order to give a more semantical latent representation.

\begin{table}[ht]
\centering
\footnotesize
\begin{adjustbox}{max width=0.48\textwidth}

\setlength{\tabcolsep}{2pt}
\renewcommand{\arraystretch}{1.15}
\begin{tabular}{|l|c|c|c|c|c|c|c|}
\hline
\multicolumn{8}{|c|}{Test on \textbf{HumanML}} \\
\hline
Trained on & Ours & Ours+HML & HML & I400 & Ours+I400 & HML+I400 & All 3 \\
\hline
MPJPE $\downarrow$     
 & 25.40 
 & \textbf{9.92}
 & \underline{\textbf{8.33}} 
 & 189.55 
 & 24.86 
 & 14.29 
 & \underline{9.98} \\
PA-MPJPE $\downarrow$  
 & 16.20 
 & \underline{6.42} 
 & \underline{\textbf{5.18}}
 & 130.52 
 & 15.88 
 & 8.92 
 & \textbf{6.11} \\
ACCL $\downarrow$     
 & 5.37 
 & \textbf{3.75}
 & \underline{\textbf{3.04}}
 & 21.04 
 & 5.49 
 & 3.91 
 & \underline{3.81} \\
\hline\hline
\multicolumn{8}{|c|}{Test on \textbf{Idea400}} \\
\hline
Train on & Ours & Ours+HML & HML & Idea400 & Ours+I400 & HML+I400 & All 3 \\
\hline
MPJPE $\downarrow$     
 & 20.87 
 & 16.63 
 & 18.92 
 & \underline{\textbf{6.52}}
 & \underline{8.89}
 & 12.61 
 & \textbf{8.77} \\
PA-MPJPE $\downarrow$  
 & 12.92 
 & 10.50 
 & 12.28 
 & \underline{\textbf{4.18}}
 & \underline{5.48} 
 & 7.52 
 & \textbf{5.25} \\
ACCL $\downarrow$     
 & 6.38 
 & 6.20 
 & 6.97 
 & \underline{\textbf{3.12}}
 & \textbf{4.41} 
 & 5.82 
 & \underline{4.97} \\
\hline\hline
\multicolumn{8}{|c|}{Test on \textbf{Ours (XmoPipe dataset)}} \\
\hline
Train on & Ours & Ours+HML & HML & Idea400 & Ours+I400 & HML+I400 & All 3 \\
\hline
MPJPE $\downarrow$     
 & 7.60 
 & \underline{\textbf{6.50}} 
 & 46.40 
 & 132.99
 & \underline{6.54}
 & 51.58 
 & \textbf{6.71} \\
PA-MPJPE $\downarrow$  
 & \textbf{4.31}
 & 4.47
 & 14.20
 & 49.63
 & \underline{4.11} 
 & 17.71 
 & \underline{\textbf{4.05}} \\
ACCL $\downarrow$     
 & 2.59 
 & \textbf{2.54} 
 & 4.05 
 & 6.93
 & \underline{\textbf{2.52}} 
 & 4.61 
 & \underline{2.64} \\
\hline
\end{tabular}
\end{adjustbox}
\caption{
Quantitative results of reconstruction on HumanML, Idea400 and XmoPipe test sets.
\underline{\textbf{Bold and underlined}} values indicate the best result, \textbf{bold} the second best, and \underline{underlined} the third best.
"Ours" denotes XmoPipe, "HML" HumanML, and "I400" Idea400.
}
\label{recons}
\end{table}


\autoref{recons} reports MPJPE, PA-MPJPE, and acceleration error (ACCL). 
As expected, models trained and evaluated on the same dataset achieve the lowest in-domain errors, and adding XmoPipe can slightly degrade this performance due to increased data variability. HumanML, a curated MoCap dataset, achieves best results due to precise MoCap capture conditions. However, the inclusion of XmoPipe also leads to more balanced performance in out-of-domain scenarios, suggesting its value for cross-dataset generalization.

In detail, training on XmoPipe alone does not always outperform MoCap datasets in in-domain reconstruction, which is consistent with the behavior of small, low-capacity models that are sensitive to noise. However, cross-dataset results highlight the complementary value of XmoPipe. Idea400 alone performs poorly on HumanML3D, while HumanML3D is not dramatically worse on Idea400, highlighting the inherent complexity of cross-domain generalization between heterogeneous motion datasets. Importantly, the best-performing configurations (excluding the trivial case of training and testing on the same dataset) often involve XmoPipe combined with another dataset. Moreover, since XmoPipe is roughly an order of magnitude larger than the other datasets and no relative weighting is applied during training, its inclusion could potentially degrade performance if the data were of low quality. Instead, we observe more stable cross-dataset behavior and improved temporal smoothness in several settings, as reflected by reduced acceleration errors. This suggests that, even for simple models, our large-scale dataset can contribute to cross-domain generalization. Finally, the dataset can be readily scaled up by adding new keywords to the collection process, allowing the acquisition of additional video data without motion capture hardware.




\subsubsection{Generation task}
\begin{figure}[h]
    \centering
    \begin{subfigure}{0.45\textwidth}
        \centering
        \includegraphics[width=\linewidth]{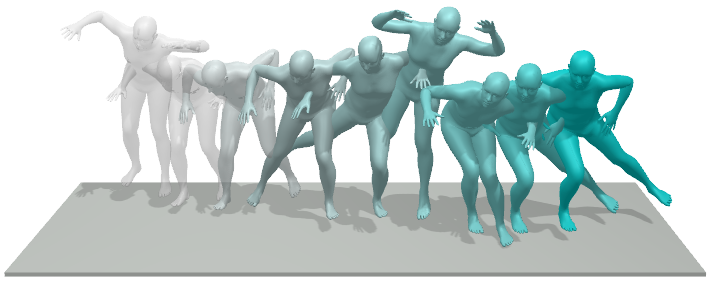}
        \caption{MDM trained on XmoPipe}
        \label{fig:img1}
    \end{subfigure}
    \hfill
    \begin{subfigure}{0.45\textwidth}
        \centering
        \includegraphics[width=\linewidth]{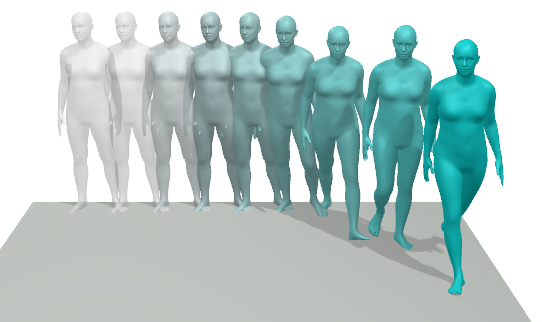}
        \caption{MDM trained on HumanML3D, the person is walking since the skiing motion does not exist in this dataset.}
        \label{fig:img2}
    \end{subfigure}

    \caption{Text-to-Motion Generation results using MDM models trained on XmoPipe vs. HumanML3D for the prompt “A person is skiing”.}
    \label{fig:comparison}
\end{figure}

We further evaluate our dataset on a more challenging and expressive task than motion reconstruction by training a Motion Diffusion Model (MDM) \cite{tevet2022humanmotiondiffusionmodel} with text–motion alignment. This task directly tests the semantic coverage and diversity of the training data, as successful generation requires capturing fine-grained action concepts beyond basic locomotion patterns. For instance, in \autoref{fig:comparison}, the model trained on XmoPipe generates motions that better match the prompt “A person is skiing” while the model trained on HumanML3D fails to produce skiing motions, defaulting to walking instead, which is expected since it does not exist on this dataset. 
While this evaluation is qualitative, it highlights failure cases that are difficult to capture with standard automatic metrics and directly reflect limitations in dataset semantic coverage.
We note that this evaluation is conducted using a single model to isolate the effect of the dataset, and extending to multiple architectures is left for future work.
Moreover, by changing the collection keywords, our pipeline can target specific motion domains, enabling scalable dataset expansion.

\subsubsection{Dataset Statistics and Diversity Analysis}

\begin{figure}[h]
    \centering
    \includegraphics[width=0.48\textwidth]{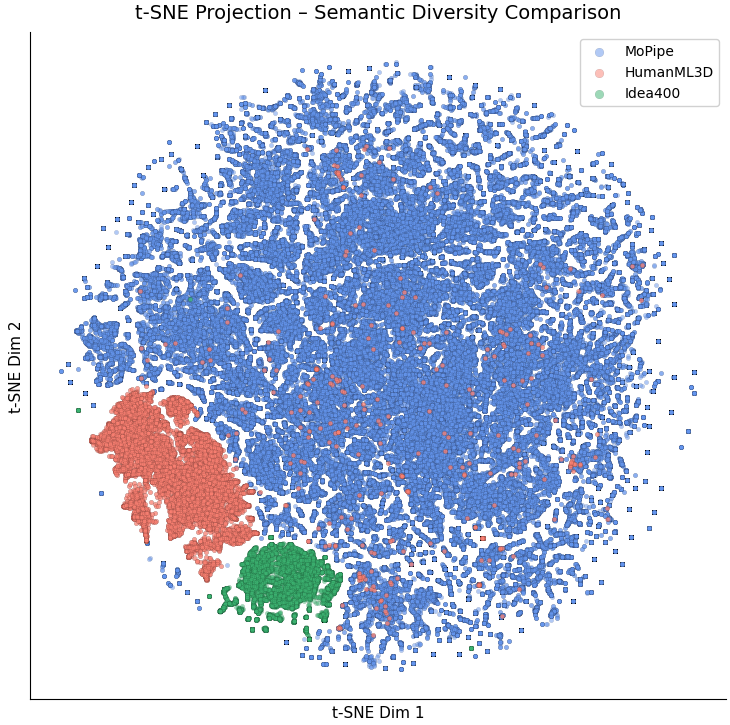}
    \caption{T-SNE of embedded action texts from each chosen datasets.}
    \label{tsne}
\end{figure}

To illustrate the diversity enabled by the extensible design of XmoPipe, 
we provide several statistical analyses of the dataset. \autoref{tsne} shows 
a t-SNE projection of sentence embeddings obtained from a sentence transformer 
applied to textual action descriptions, highlighting broad semantic diversity 
across action categories. \autoref{fig:stats} further characterizes the dataset 
through wordclouds of the four description fields (Action, Body, Scene, Style) 
and the distribution of persons per scene, revealing rich lexical variety and 
a mix of single- and multi-person setups.

Although XmoPipe contains multi-person interactions and expressive facial motions, our current evaluation focuses on single-person motion modeling to isolate the impact of large-scale data diversity. We nevertheless release the full facial motion and annotations to encourage future research on interaction-aware and facially expressive motion modeling.

\section{Conclusion and Future Work}
\label{sec:conclusion}
In this work, we introduced XmoPipe, a modular and extensible pipeline for constructing semantically enriched, multi-person human motion datasets from in-the-wild videos. As detailed in \autoref{sec:xmo}, XmoPipe combines video filtering, monocular body and facial reconstruction of multi-person, data fusion, temporal smoothing, and automatic semantic annotation to transform raw web videos into structured SMPL-X motion data paired with rich textual descriptions. We demonstrated that the resulting dataset is directly suitable for downstream learning, notably by training Motion Diffusion Models (MDM), where increased motion and semantic diversity lead to more diverse text-to-motion generation.

More broadly, although in-the-wild monocular videos and automatic annotations inevitably introduce noise, our results suggest that large-scale automatically collected data can complement MoCap datasets and improve cross-dataset generalization. The pipeline can be easily extended by introducing new collection keywords, enabling rapid expansion of the dataset or the creation of domain-specific corpora (e.g., sports- or art-focused motion datasets) with minimal manual intervention.

Although our experiments mainly focus on single-person motion modeling, XmoPipe produces data that is compatible with multi-person and expressive motion generation settings. An important future direction is to evaluate our dataset with generative models designed for these scenarios, such as InterGen\cite{liang2024intergen} for two-person motion generation and EMAGE\cite{liu2024emageunifiedholisticcospeech} for motion with facial expressions. Such evaluations would further assess the suitability of automatically collected in-the-wild data for such generation tasks.
In addition, the pipeline itself can be extended to extract richer motion representations. In particular, hand pose reconstruction could be incorporated using models such as WiLoR~\cite{potamias2025wilorendtoend3dhand}, thereby completing the full SMPL-X representation, although this remains challenging in fully in-the-wild settings.

\newpage
\clearpage
\section*{Acknowledgments}

This work was granted access to the HPC resources of IDRIS under the allocation 2024-AD011015875 and 2025-AD011015875R1 granted by GENCI on the V100 partition of the Jean Zay supercomputer.


\bibliographystyle{unsrt}
{\footnotesize
\bibliography{main}
}
\appendix
\section{Supplementary material}
\label{sec:annex}
Yolov11n-pose confidence threshold to filter the frames : \textbf{0.65}\\
Vitpose-h model used to get the body keypoints for GVHMR: \textbf{ViTPose\_huge\_coco\_256x192} \\
Vitpose-h confidence thresholds for GVHMR: \textbf{0.55} as a mean confidence, \textbf{0.5} to flag as unsure\\
\begin{math}\sigma=1\end{math} for the gaussian filter on temporal dimension.
We filter out face data when there are less than \textbf{5} frames in a row detected

\begin{table}[h]
\centering
\label{tab:vae_hparams}
\resizebox{0.48\textwidth}{!}{

\begin{tabular}{| l | c |}
\hline
\textbf{Parameter} & \textbf{Value} \\
\hline
Architecture & Sequential VAE\\
Encoder / Decoder & CNN with ResNet blocks \\
Downsampling factor & 8 \\
Latent dimension & (8*) 256 \\
Datasets & XmoPipe, HumanML3D, Idea400\\
Learning rate & $1 \times 10^{-5}$ \\
Batch size & 256 \\
Motion size & [64,263] \\
Trained for & $10^{6}$ steps \\
\hline
\end{tabular}
}
\caption{Sequential VAE training and architecture hyperparameters}
\end{table}

\end{document}